\documentclass{article}
\usepackage{spconf,amsmath,epsfig}
\usepackage{ctable}
\usepackage{subcaption}
\usepackage{color}
\usepackage{multirow}

\definecolor{myred}{RGB}{255, 0, 0}
\definecolor{mygreen}{RGB}{0, 165, 60}
\definecolor{myblue}{RGB}{0, 112, 192}

\title{CROSS-DOMAIN CNN FOR HYPERSPECTRAL IMAGE CLASSIFICATION}
\name{
Hyungtae Lee$^{\dagger\ddagger}$,
Sungmin Eum$^{\dagger\ddagger}$,
Heesung Kwon$^{\ddagger}$
}

\address{$^{\dagger}$Booz Allen Hamilton Inc., McLean, Virginia, U.S.A.\\
$^{\ddagger}$U.S. Army Research Laboratory, Adelphi, Maryland, U.S.A.\\
{\small {\tt lee\_hyungtae@bah.com, eum\_sungmin@bah.com, heesung.kwon.civ@mail.mil}}}

\begin{document}
%
\maketitle
\begin{abstract}
In this paper, we address the dataset scarcity issue with the hyperspectral image classification. As only a few thousands of pixels are available for training, it is difficult to effectively learn high-capacity Convolutional Neural Networks (CNNs). To cope with this problem, we propose a novel cross-domain CNN containing the shared parameters which can co-learn across multiple hyperspectral datasets. The network also contains the non-shared portions designed to handle the dataset-specific spectral characteristics and the associated classification tasks. Our approach is the first attempt to learn a CNN for multiple hyperspectral datasets, in an end-to-end fashion. Moreover, we have experimentally shown that the proposed network trained on three of the widely used datasets outperform all the baseline networks which are trained on single dataset.
\end{abstract}
\section{Introduction}
\label{sec:intro}

The introduction of convolutional neural network (CNN) has brought forth unprecedented performance increase for classification problems in many different domains including RGB, RGBD, and hyperspectral images.~\cite{SGuptaECCV2014,RGirshickTPAMI2016,KHeCVPR2016,HLeeIGARSS2016,HLeeTIP2017} Such performance increase was made possible due to the ability of CNN being able to learn and express the deep and wide connection between the input and the output using a huge number of parameters. In order to learn such a huge set of parameters, having a large scale dataset has become a significant requirement. When the size of the given dataset is insufficient to learn a network, one may consider using a larger external dataset to better learn the large set of parameters. For instance, Girshick et al. \cite{RGirshickTPAMI2016} introduced a domain adaptation approach where the network is trained on a large scale source domain (ImageNet dataset~\cite{JDengCVPR2009}) and then finetuned on a target domain (object detection dataset~\cite{MEveringhamIJCV2010}).

When applying CNN to hyperspectral image classification problem, we also face the similar issue as there are no large scale hyperspectral dataset available. A typical hyperspectral image classification data only contains between 10k and 100k pixels where very small portion of those pixels are being used for training. In order to tackle such data scarcity issue, we need a way to make use of multiple hyperspectral dataset (domain).

There are several challenges that arise when devising an approach which can be applied to multiple hyperspectral domains. First of all, all the hyperspectral datasets contain different wavelength range and spectral reflectance bands. Furthermore, applying a domain adaptation approach \cite{RGirshickTPAMI2016} is infeasible as a large scale auxiliary dataset for the hyperspectral image classification is not available.

\begin{figure}[t]
\begin{center}
\centerline{\includegraphics[width=\linewidth,trim=0mm 60mm 7mm 7mm]{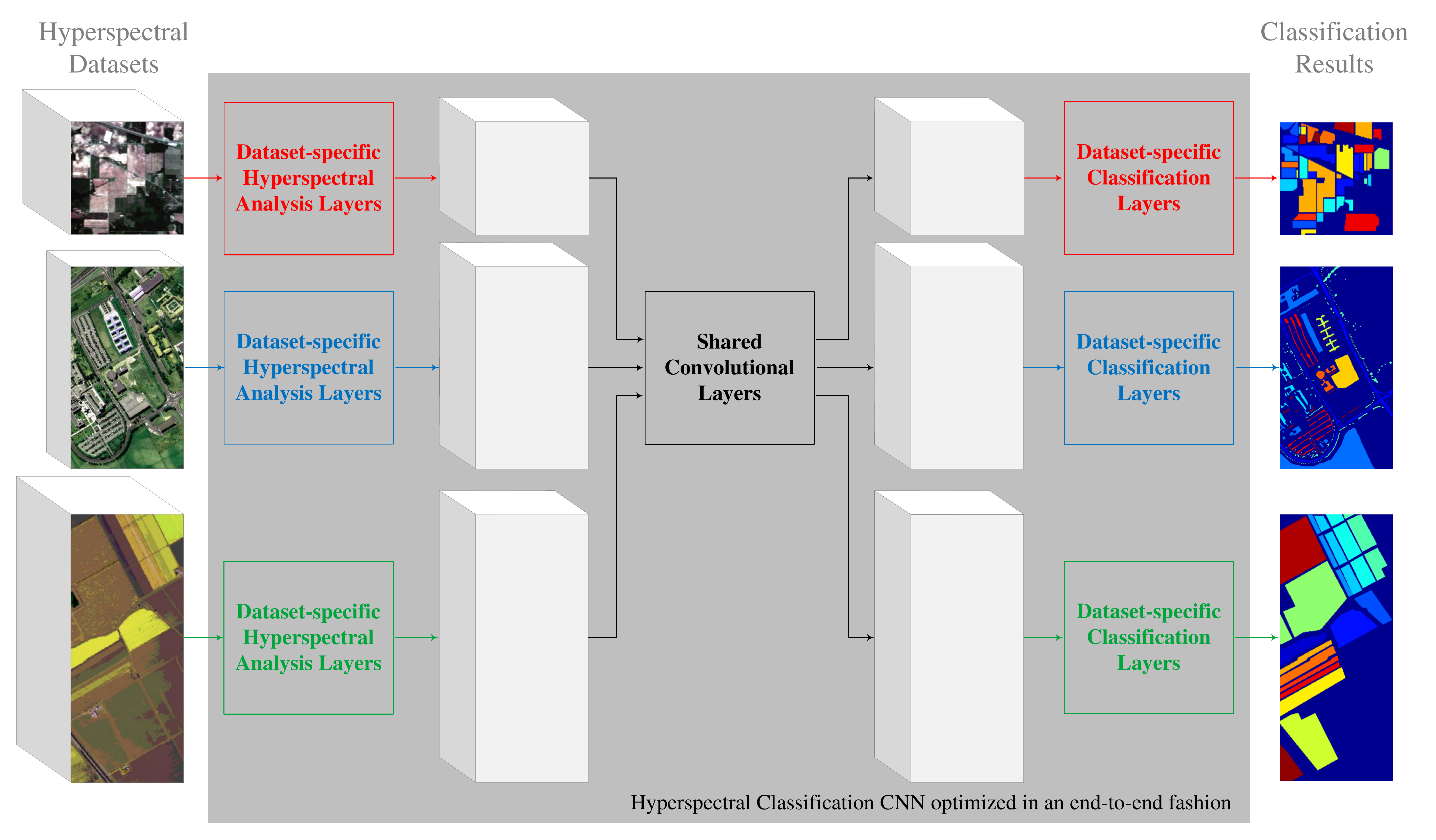}}
\end{center}
\caption{{\bf Cross-Domain CNN.} While the non-shared portions of the CNN (``dataset-specific hyperspectral analysis layers'' and ``dataset-specific classification layers'') are handling the dataset-specific classification tasks, the ``shared convolutional layers'' generate common information which applies to all the datasets. Note that both shared and non-shared portions of the CNN are optimized in an end-to-end fashion.}
\label{fig:concept}
\end{figure}

Therefore, we have developed a novel cross-domain CNN architecture which simultaneously performs network learning and classification on multiple hyperspectral datasets. The architecture consists of three components: dataset-specific hyperspectral analysis layers, shared convolutional layers, and dataset-specific classification layers. In the front-end portion, ``dataset-specific hyperspectral analysis layers'' are present to analyze the spatial-spectral information. The back-end is built with the ``dataset-specific classification layers'' which performs the classification for different dataset. These two components are connected by the ``shared convolutional layers'' which learns the common information across the domains. All three components are being optimized in an end-to-end fashion. The information acquired from multiple datasets is fed through the layers concurrently during training, which leads to the better learning of the shared convolutional layers via dataset augmentation. The overall CNN architecture is depicted in Figure \ref{fig:concept}.

In this paper, we have used the three mostly used hyperspectral datasets (Indian Pines, Salinas, and University of Pavia) to demonstrate the effectiveness of having a cross-domain CNN. The experimental results show that our novel architecture outperforms the baseline networks (i.e., only one dataset used to train the network) by about 1.5\% to 3\% consistently in terms of classification accuracy.

The contributions of our approach are listed as below.
\begin{enumerate}
    \item First attempt to learn a CNN with multiple hyperspectral datasets.
    \item A novel cross-domain CNN optimized in an end-to-end fashion.
    \item Consistent classification accuracy increase on all datasets.
\end{enumerate}

\section{The Proposed Method}
\label{sec:proposed_method}

\subsection{Architecture}

\begin{figure}[t]
\begin{center}
\centerline{\includegraphics[width=\linewidth,trim=7mm 115mm 7mm 0mm]{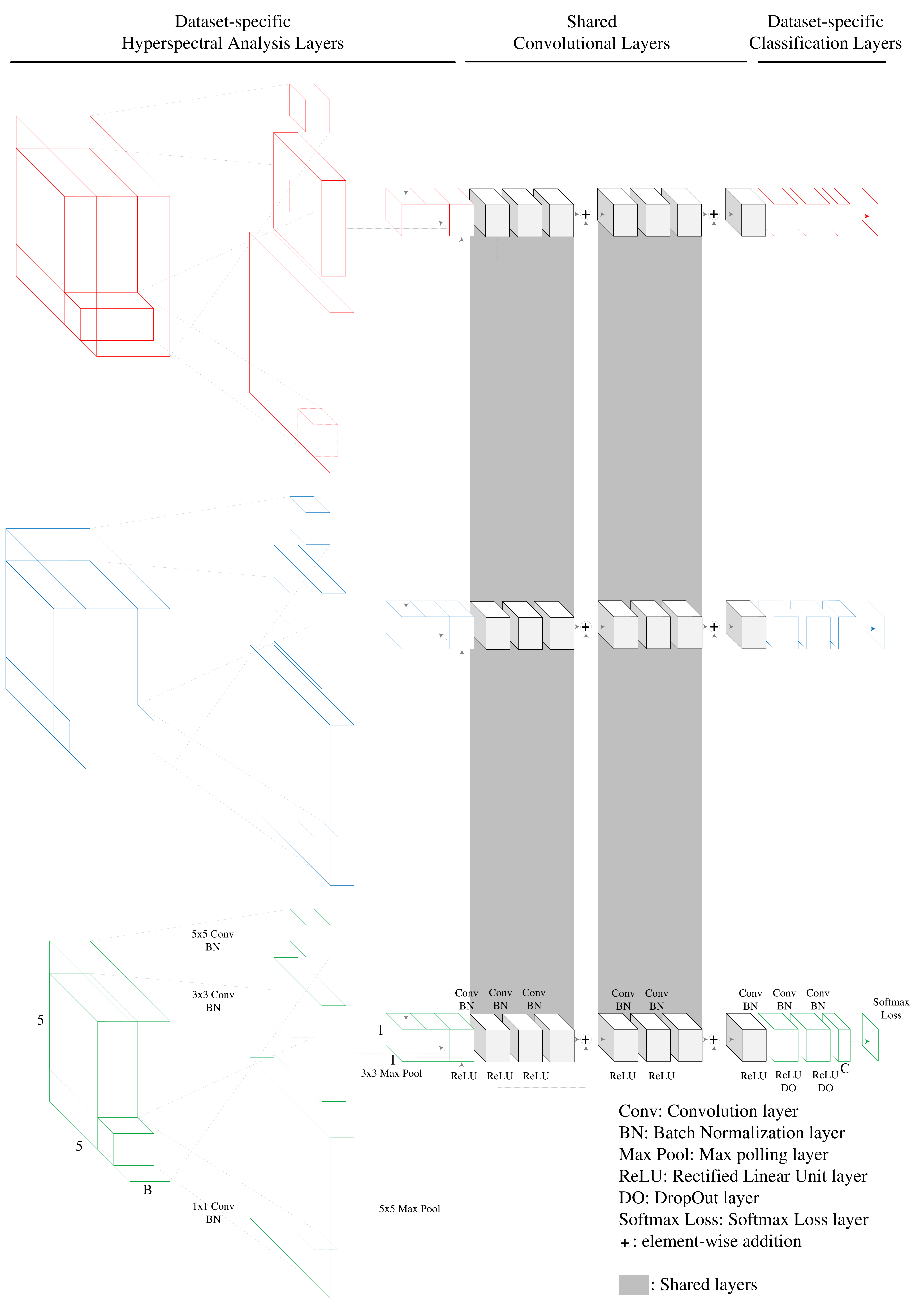}}
\end{center}
\caption{{\bf Architecture.} \textcolor{myred}{Red}, \textcolor{myblue}{blue}, \textcolor{mygreen}{green} blocks indicate the non-shared portion of the CNN while the black ones are shared across all the streams for different datasets. Detailed specifications on the architecture are denoted on the last (\textcolor{mygreen}{green}) stream. {\bf B} and {\bf C} indicate the number of spectral bands and the number of classes, respectively, which are different for all the datasets.}
\label{fig:flowchart}
\end{figure}

Figure~\ref{fig:flowchart} shows the proposed cross-domain CNN architecture.\\

\noindent{\bf Backbone CNN architecture.} For each dataset-specific stream in the proposed architecture, we have used the modified version of the 9-layered hyperspectral image classification CNN introduced by Lee and Kwon~\cite{HLeeIGARSS2016,HLeeTIP2017}. The modification has been carried out by adding a batch normalization (BN) layer~\cite{SIoffeICML2015} after each convolutional layer while removing all the local response normalization layers. The BN layer computes the mean and the variance of the samples in each mini-batch and performs normalization, that is, it fits the samples in the mini-batch to a normal distribution. By introducing the BN, we can bypass the process of data normalization for both training and testing. In addition, bias terms are no longer required.

The backbone CNN is a fully convolutional network and contains the sequentially connected multi-scale filter bank, one convolutional layer, two residual modules and three convolutional layers. Each multi-scale filter bank~\cite{CSzegedyCVPR2015} consists of 1$\times$1, 3$\times$3, and 5$\times$5 filters to analyze the spatial-spectral characteristics. Each residual module~\cite{KHeCVPR2016} includes two convolutional layers. The residual modules allow ease of learning of the deep network. To the best of our knowledge, this architecture is the first attempt to go deeper than three layers by adopting the residual modules within the domain of hyperspectral image classification.\\

\begin{table}
\caption{{\bf Specifications for different hyperspectral datasets.} Reduced bands are acquired by removing the bands which correspond to the water absorption. Note that, datasets acquired using the same type of sensor can have different reduced bands.}
\label{tab:domains}
\setlength{\tabcolsep}{2.3pt}
\renewcommand{\arraystretch}{1.4}
{\footnotesize
\begin{tabular}{l|c|c|c|c}
\specialrule{.15em}{.05em}{.05em} 
Dataset & Sensor & Range & Bands & Reduced Bands \\\specialrule{.15em}{.05em}{.05em}
Indian Pines & \multirow{2}{*}{AVIRIS} & \multirow{2}{*}{0.4$\mu$m $\sim$ 2.5$\mu$m} & \multirow{2}{*}{224} & 200\\\cline{1-1}\cline{5-5}
Salinas & & & & 204 \\\hline
University of Pavia & ROSIS & 0.43$\mu$m $\sim$ 0.86$\mu$m & 115 & 103 \\\specialrule{.15em}{.05em}{.05em} 
\end{tabular}
}
\end{table}

\noindent{\bf Cross-domain CNN architecture.} To perform an RGB-to-RGB domain adaptation process using CNNs~\cite{RGirshickTPAMI2016,MOquabCVPR2014}, we can follow a traditional approach of replacing the classification layers to fit the target dataset. However, when source and target domains are hyperspectral (where each dataset carries its unique characteristics as shown in Table \ref{tab:domains}), simply replacing the classification layer does not work physically. Because of this, we have devised dataset-specific layers not only at the latter portion of the network, but also before the shared convolutional layers to adaptively intake the different datasets.

The multi-scale filter bank in each dataset-specific stream, which is responsible for analyzing the spatial-spectral characteristics, is assigned as the ``dataset-specific hyperspectral analysis layers''. The last three convolutional layers function as the ``dataset-specific classification layers''. The remaining layers in the middle of the architecture which consist of the second convolutional layer and two residual modules are assigned as the cross-domain ``shared convolution layers''.

\subsection{Optimization}

Each layer consists of 128 convolutional filters. The first, second, and ninth convolutional layers are initialized according to Gaussian distribution with mean of 0 and standard deviation of 0.01. Remaining convolutional layers are initialized with zero-mean Gaussian and standard deviation of 0.005. To provide richer set of samples and to avoid over-fitting, the training samples are augmented eight-fold by mirroring each sample across the vertical, horizontal, and two diagonal axes.

We have used stochastic gradient descent (SGD) to train the network with a batch size of 10 samples. The training process was initiated with the learning rate of 0.001 and iterated 100k times with a step size of 40k. We set the momentum, gamma, and weight decay as 0.9, 0.1, and 0.0005, respectively.

When learning the ``shared convolutional layers'', we multiply $1/N$ (where $N$ is the number of domains involved in the training process) to the base learning rate because updating the weights in these layers are affected by all $N$ domain-specific networks when back-propagation takes place at each iteration. In our case, $N$ is set to be 3 as we have used 3 different datasets.

\section{Evaluation}
\label{sec:eval}

\begin{table*}
\caption{Selected classes for evaluation and the numbers of training and test samples.}
\label{tab:dataset_info}
\begin{tabular}{ccc}
\begin{subfigure}{0.29\textwidth}
\caption{Indian Pines}
\setlength{\tabcolsep}{9.5pt}
\renewcommand{\arraystretch}{1.1}
{\footnotesize
\begin{tabular}{lcc}
\specialrule{.15em}{.05em}{.05em} 
Class & Training & Test \\\specialrule{.15em}{.05em}{.05em} 
Corn-notill & 200 & 1228 \\
Corn-mintill & 200 & 630 \\
Grass-pasture & 200 & 283 \\
Hay-windrowed & 200 & 278 \\
Soybean-notill & 200 & 772 \\
Soybean-mintill & 200 & 2255 \\
Soybean-clean & 200 & 393 \\
Woods & 200 & 1065 \\\specialrule{.15em}{.05em}{.05em} 
Total & 1600 & 6904 \\\specialrule{.15em}{.05em}{.05em} 
\end{tabular}
}
\vspace{8.3em}
\end{subfigure}
&
\begin{subfigure}{0.37\textwidth}
\caption{Salinas}
\setlength{\tabcolsep}{9.5pt}
\renewcommand{\arraystretch}{1.1}
{\footnotesize
\begin{tabular}{lcc}
\specialrule{.15em}{.05em}{.05em} 
Class & Training & Test \\\specialrule{.15em}{.05em}{.05em} 
Brocooli green weeds 1 & 200 & 1809 \\
Brocooli green weeds 2 & 200 & 3526 \\
Fallow & 200 & 1776 \\
Fallow rough plow & 200 & 1194 \\
Fallow smooth & 200 & 2478 \\
Stubble & 200 & 3759 \\
Celery & 200 & 3379 \\
Grapes untrained & 200 & 11071 \\
Soil vineyard develop & 200 & 6003 \\
Corn senesced green weeds & 200 & 3078 \\
Lettuce romaines, 4 wk & 200 & 868 \\
Lettuce romaines, 5 wk & 200 & 1727 \\
Lettuce romaines, 6 wk & 200 & 716 \\
Lettuce romaines, 7 wk & 200 & 870 \\
Vineyard untrained & 200 & 7068 \\
Vineyard vertical trellis & 200 & 1607 \\\specialrule{.15em}{.05em}{.05em} 
Total & 3200 & 50929 \\\specialrule{.15em}{.05em}{.05em} 
\end{tabular}
}
\end{subfigure}
&
\begin{subfigure}{0.28\textwidth}
\caption{University of Pavia}
\setlength{\tabcolsep}{9.5pt}
\renewcommand{\arraystretch}{1.1}
{\footnotesize
\begin{tabular}{lcc}
\specialrule{.15em}{.05em}{.05em} 
Class & Training & Test \\\specialrule{.15em}{.05em}{.05em} 
Asphalt & 200 & 6431 \\
Meadows & 200 & 18449 \\
Gravel & 200 & 1899 \\
Trees & 200 & 2864 \\
Sheets & 200 & 1145 \\
Bare soils & 200 & 4829 \\
Bitumen & 200 & 1130 \\
Bricks & 200 & 2482 \\
Shadows & 200 & 747 \\\specialrule{.15em}{.05em}{.05em} 
Total & 1800 & 40976 \\\specialrule{.15em}{.05em}{.05em} 
\end{tabular}
}
\vspace{7.5em}
\end{subfigure}
\end{tabular}
\end{table*}

\subsection{Evaluation Settings}

We have used three hyperspectral datasets (Indian Pines, Salinas, and University of Pavia) for the experiments. The Indian Pines dataset includes 145$\times$145 pixels and 200 spectral reflectance bands which cover the range from 0.4 to 2.5 $\mu$m with a spatial resolution of 20 m. The Indian Pines dataset has 16 classes but only 8 classes with relatively large numbers of samples are used. The Salinas dataset contains 16 classes with 512$\times$217 pixels and 204 spectral bands and a high spatial resolution of 3.7 m. Salinas dataset shares the frequency characteristics with the Indian Pines dataset as the same sensor (AVIRIS) was used for the data acquisition. The University of Pavia dataset, which was acquired using ROSIS sensor, provides 9 classes and 610$\times$340 pixels with 103 spectral bands which cover the spectral range from 0.43 to 0.86 $\mu$m with a spatial resolution of 1.3 m. For the Salinas dataset and the University of Pavia dataset, we use all available classes because both datasets do not contain classes with a relatively small number of samples. Each dataset has been randomly partitioned into train and test sets according to Table~\ref{tab:dataset_info}.

\subsection{Performances}

\noindent{\bf Classification accuracy.} As shown in Table~\ref{tab:accuracy}, the proposed cross-domain CNN outperforms the individual networks on the Indian Pines, Salinas, and University of Pavia by 1.5\%, 1.5\%, and 3.2\%, respectively. Note that the individual networks have been trained without the shared portion of the network. We observe that using a bigger and richer dataset (i.e., using all three datasets to learn one single network) in training the cross-domain CNN assisted in boosting up the performance.\\

\begin{table}
\caption{Classification accuracy.}
\label{tab:accuracy}
\setlength{\tabcolsep}{6.0pt}
\renewcommand{\arraystretch}{1.4}
{\footnotesize
\begin{tabular}{l|c|c|c}
\specialrule{.15em}{.05em}{.05em} 
Dataset & Individual CNN & Cross-Domain CNN & Gain \\\specialrule{.15em}{.05em}{.05em}
Indian Pines & .907 & {\bf .922} & +.015 \\
Salinas & .893 & {\bf .908} & +.015 \\
University of Pavia & .921 & {\bf .953} & +.032 \\\specialrule{.15em}{.05em}{.05em} 
\end{tabular}
}
\end{table}

\begin{figure}[t]
\begin{center}
\centerline{\includegraphics[width=\linewidth,trim=5mm 130mm 5mm 105mm]{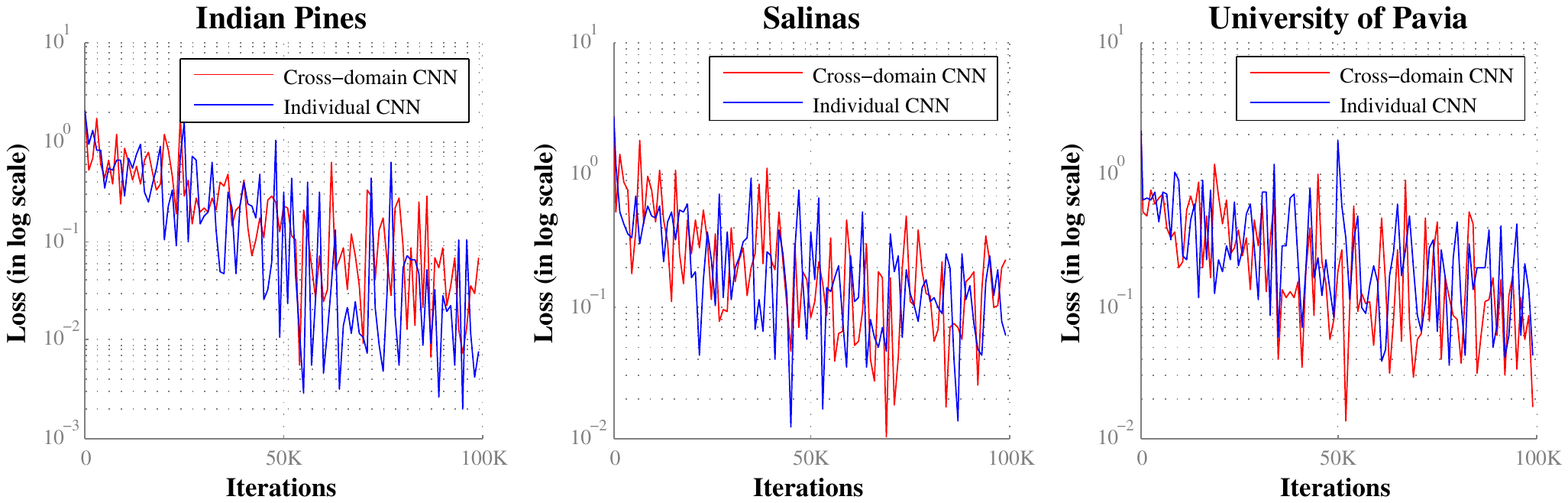}}
\end{center}
\caption{Comparing training loss evolution: Cross-domain CNN vs. Individual CNNs.}
\label{fig:loss}
\end{figure}

\noindent{\bf Training loss analysis.}
As shown in Figure~\ref{fig:loss}, training loss evolution for the cross-domain CNN does not show much difference when compared with the cases where CNNs are trained separately for different datasets. However, we observe that the cross-domain CNN shows better performance in terms of classification accuracy. This indicates that the performance gain achieved by the cross-domain CNN is due to the effort in addressing the overfitting issue. As we have used a larger dataset (i.e., combined set of all three) to train one network, discrepancies between the training and the test sets are decreased.\\


\noindent{\bf Test time.} We have evaluated the computation time of our cross-domain CNN when three images (one from each dataset) are fed in for testing. It takes 0.5875 seconds for processing all three images when tested on NVIDIA TITAN XP GPU.\\

\section{Conclusion}
\label{conclusion}
In this paper, we have introduced a novel cross-domain CNN which can concurrently learn and perform the hyperspectral image classification for multiple datasets. As the shared portion of our network is being trained using multiple hyperspectral datasets, the proposed approach is more effective in optimizing the high capacity CNN than the cases where only a single dataset (domain) is being used. Our approach is the first attempt to exploit multiple hyperspectral datasets for training a CNN in an end-to-end fashion. We have experimentally demonstrated that using the shared layers across the domains brings notable classification accuracy improvements when compared to the individually trained cases.

\bibliographystyle{IEEE}
\bibliography{refs}

\end{document}